# Bayer Demosaicking Using Optimized Mean Curvature over RGB channels

Rui Chen, Huizhu Jia, Xiange Wen and Xiaodong Xie

Color artifacts of demosaicked images are often found at contours due to interpolation across edges and cross-channel aliasing. To tackle this problem, we propose a novel demosaicking method to reliably reconstruct color channels of a Bayer image based on two different optimized mean-curvature (MC) models. The missing pixel values in green (G) channel are first estimated by minimizing a variational MC model. The curvatures of restored G-image surface are approximated as a linear MC model which guides the initial reconstruction of red (R) and blue (B) channels. Then a refinement process is performed to interpolate accurate full-resolution R and B images. Experiments on benchmark images have testified to the superiority of the proposed method in terms of both the objective and subjective quality.

*Introduction:* Demosaicking is a technique for reconstructing a full-color image from the raw image captured by a digital color camera that utilizes a single image sensor with a color filter array (CFA). For the popular Bayer CFA pattern, only one color component is captured at each pixel and the other missing components must be estimated to produce high-quality RGB images. A number of Bayer demosaicking approaches have been developed by mainly exploiting the spatial and spectral characteristics of raw images [1-6]. The high interchannel and intrachannel correlations are often assumed to restore the unknown color components. However, for images with sharp color transition and high color saturation, the demosaicking performance is subject to the unreliable estimate of local image structures and the extent of spectral correlations. This Letter proposes a novel demosaicking algorithm by transforming into a curvature domain to reconstruct full-resolution RGB channels, which can improve the structure-preserving capability, enhance the weak correlations and reduce the color artifacts. The whole demosaicking process is shown in Fig. 1.

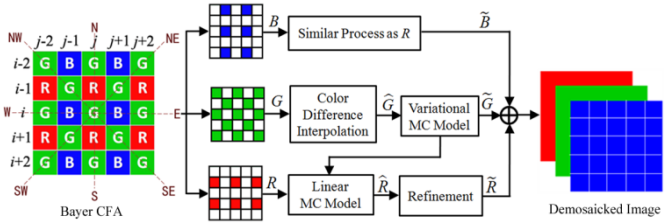

**Fig. 1** *Flowchart of the proposed demosaicking method*

*Green Channel Reconstruction:* Because original G channel is sampled twice more than the other two channels and preserves much more image structural information, the subsampled image $G$ is first interpolated to estimate an initial full-resolution image $\hat{G}$ by utilizing an arbitrary color difference interpolation algorithm. We here adopt a current state-of-the-art algorithm, the gradient based threshold free (GBTF) [3], to obtain the missing G pixel values as

$$\begin{cases} \hat{G}_r = R(i,j) + \Delta_{g,r}(i,j) \\ \hat{G}_b = B(i,j) + \Delta_{g,b}(i,j) \end{cases} \quad (1)$$

where the pixel values, $\hat{G}_r$ and $\hat{G}_b$, at the R and B pixel locations are obtained by adding estimated color differences $\Delta_{g,r}$ and $\Delta_{g,b}$ to R and B pixel values, respectively. The inaccurate differences often cause the artifacts.

To prevent possible artifacts and preserve detailed image structures, we use mean curvature of the associated image surface as a good geometric metrics to characterize edges and textures. The mean curvature $\kappa$ can be simplified along level lines passing through smooth surface [7]. Then we define it as:

$$\kappa(I) = \mathbf{div}\left(\frac{\nabla I}{\sqrt{|\nabla I|^2 + 1}}\right) \quad (2)$$

Here $\nabla$ denotes the gradient operator of the image $I$ and **div** is the divergence operator. This realization form of mean curvature can effectively hold geometric features of image surface and differentiate with non-regularity structures. By treating the interpolation error $\varepsilon$ or artifacts of Eq. (1) as additive noise, we model the expected G-channel image as $\tilde{G} = \hat{G} + \varepsilon$. The $\ell_1$ norm of curvature $\kappa$ as the regularizer allows jumps of image structures as well as continuous intensity distributions along the contours while minimizing absolute value of $\kappa$ in piecewise smooth regions. These properties suggest that the artifact parts of images can be removed while edges and textures can be preserved. Accordingly, the proposed variational MC model is written as follows:

$$\tilde{G} = \arg\min \frac{1}{2}\int_\Omega (\hat{G} - \tilde{G})^2 + \lambda \int_\Omega |\kappa(\tilde{G})| \quad (3)$$

To reconstruct final G-channel $\tilde{G} \in BV(\Omega)$, we use augmented Lagrangian method to solve optimization problem of the Eq. (3). Let's first introduce the auxiliary variables: $q = \mathbf{div}\,n$, $n = p/|p| \in R^3$, $p = [\nabla \tilde{G}, 1] \in R^3$, $m \in R^3$. The associated augmented Lagrangian functional reads as follows:

$$\begin{aligned}
&\mathcal{L}(\tilde{G}, q, p, n, m; \lambda_1, \lambda_2, \lambda_3, \lambda_4) \\
&= \frac{1}{2}\int_\Omega (\hat{G}-\tilde{G})^2 + \lambda\int_\Omega |q| + r_1 \int_\Omega (|p| - p \cdot m) + \lambda_1 \int_\Omega (|p| - p \cdot m) \\
&\quad + \frac{r_2}{2}\int_\Omega |p - [\nabla \tilde{G}, 1]|^2 + \int_\Omega \lambda_2 \cdot (p - [\nabla \tilde{G}, 1]) + \frac{r_3}{2}\int_\Omega (q - \mathbf{div}\,n)^2 \\
&\quad + \lambda_3 \int_\Omega (q - \mathbf{div}\,n) + \frac{r_4}{2}\int_\Omega |n - m|^2 + \int_\Omega \lambda_4 \cdot (n - m) + \delta_\Gamma(m)
\end{aligned} \quad (4)$$

where $\lambda$ is the regularization parameter. $\lambda_1, \lambda_3 \in R$ and $\lambda_2, \lambda_4 \in R^3$ are Lagrange multipliers. $r_1, r_2, r_3, r_4 \in R$ are step parameters. The variable $m$ is required to lie in the set $\Gamma$. $\delta_\Gamma(\cdot)$ is the pulse function on $\Gamma$. The subproblems associated with each variable are derived to find the optimal solution of Eq. (4). For $\tilde{G}$–subproblem, the minimizer is determined by associate Euler-Lagrange equation. We here employ Gauss-Seidel method to solve this equation and get

$$\begin{cases}(1+\frac{4r_2}{h^2})\tilde{G}^{k+1}_{(i,j)} = g_{(i,j)} + \frac{r_2}{h^2}[\tilde{G}^{k+1}_{(i-1,j)} + \tilde{G}^{k}_{(i+1,j)} + \tilde{G}^{k+1}_{(i,j-1)} + \tilde{G}^{k}_{(i,j+1)}] \\ g_{(i,j)} = \hat{G} - \partial_x(r_2 p_1 + \lambda_{21}) - \partial_y(r_2 p_2 + \lambda_{22}) \end{cases} \quad (5)$$

where $p = [p_1, p_2, p_3]$ and $\lambda_2 = [\lambda_{21}, \lambda_{22}, \lambda_{23}]$. $\tilde{G}^{k+1}_{(i,j)}$ denotes a updated value of $k+1$-th iteration at the pixel point $(i, j)$. $\partial_x$ and $\partial_y$ denote differential operators. $h$ is the discretized spatial size. In a similar way, the variable $n$ is updated. Other subproblems can be solved using Newton method [8].

*Red and Blue Channel Reconstruction:* The obtained $\tilde{G}$ is used to guide the reconstruction of the missing R and B channels. The high accuracy of G values can facilitate the color interpolation in R and B spaces. Based on the observation that the inter-channel curvature profiles have higher correlation than the usual differences or ratios of color components, we use the mean curvature measured from full-resolution image $\tilde{G}$ as a guided mask, which can be further approximated into a linear MC model. For the $(i,j)$th pixel of the image $\tilde{G}$, we first compute second-order finite difference approximations of $|\nabla \tilde{G}|$:

$$\begin{cases} d_{ij,W} = [(\tilde{G}_{(i,j)} - \tilde{G}_{(i-1,j)})^2 + (\tilde{G}_{(i-1,j+1)} + \tilde{G}_{(i,j+1)} - \tilde{G}_{(i-1,j-1)} - \tilde{G}_{(i,j-1)})^2/16]^{1/2} \\ d_{ij,S} = [(\tilde{G}_{(i,j)} - \tilde{G}_{(i,j-1)})^2 + (\tilde{G}_{(i+1,j)} + \tilde{G}_{(i+1,j-1)} - \tilde{G}_{(i-1,j)} - \tilde{G}_{(i-1,j-1)})^2/16]^{1/2} \\ d_{ij,E} = d_{ij,W} \;\; ;\;\; d_{ij,N} = d_{ij,S} \end{cases} \quad (6)$$

So far now we approximate the curvature $\kappa$ in Equation (2) by a combination of orthogonal difference projections in *x*- and *y*-coordinate directions. To smooth local extrema, the gradient magnitude $|\nabla \tilde{G}|$ is weighted the directional curvature terms. Then it follows from Eq. (6) and a linear MC model can get

$$\begin{aligned}\kappa(\tilde{G}_{(i,j)}) &\simeq |\nabla \tilde{G}_{(i,j)}|\left(\frac{\Delta_x \tilde{G}_{(i,j)}}{|\nabla \tilde{G}_{(i,j)}|}\right)_x + |\nabla \tilde{G}_{(i,j)}|\left(\frac{\Delta_y \tilde{G}_{(i,j)}}{|\nabla \tilde{G}_{(i,j)}|}\right)_y \\ &\simeq u_{ij,W}\tilde{G}_{(i-1,j)} + u_{ij,E}\tilde{G}_{(i+1,j)} + u_{ij,S}\tilde{G}_{(i,j-1)} + u_{ij,N}\tilde{G}_{(i,j)} - 4 \end{aligned} \quad (7)$$

where $\Delta_x$ and $\Delta_y$ denote the forward differences. $u_{ij,W}, u_{ij,E}, u_{ij,S}$ and $u_{ij,N}$ have

$$u_{ij,W} = \frac{2d_{ij,E}}{d_{ij,W}+d_{ij,E}}; u_{ij,E} = \frac{2d_{ij,W}}{d_{ij,W}+d_{ij,E}}; u_{ij,S} = \frac{2d_{ij,N}}{d_{ij,S}+d_{ij,N}}; u_{ij,N} = \frac{2d_{ij,S}}{d_{ij,S}+d_{ij,N}} \quad (8)$$

The weighted coefficients in model (7) can be computed from image $\tilde{G}$ based on Eq. (8). Since these curvature-related coefficients characterize local image structures in RGB channels, the missing R components are reliably estimated using a set of same coefficients, which measure the importance of neighboring pixels of whose larger values are assigned to pixels at edges than those in flat areas. We consider that R channel is interpolated in color difference domain because the interpolation performance is generally affected by the smoothness degree of local structures. The missing component $R_{(i,j)}$ at B sampling position is initially interpolated using four color differences $\{\zeta_{ij,W}, \zeta_{ij,E}, \zeta_{ij,S}, \zeta_{ij,N}\}$. Then the estimated R component $\hat{R}_{(i,j)}$ is interpolated as

$$\begin{cases} \hat{R}_{(i,j)} = \tilde{G}_{(i,j)} + \dfrac{u_{ij,W} \cdot \zeta_{ij,W} + u_{ij,E} \cdot \zeta_{ij,E} + u_{ij,S} \cdot \zeta_{ij,S} + u_{ij,N} \cdot \zeta_{ij,N}}{u_{ij,W} + u_{ij,E} + u_{ij,S} + u_{ij,N}} \\ \zeta_{ij,W} = R_{(i-1,j-1)} - \tilde{G}_{(i-1,j-1)} \,;\, \zeta_{ij,E} = R_{(i+1,j+1)} - \tilde{G}_{(i+1,j+1)} \\ \zeta_{ij,S} = R_{(i+1,j-1)} - \tilde{G}_{(i+1,j-1)} \,;\, \zeta_{ij,N} = R_{(i-1,j+1)} - \tilde{G}_{(i-1,j+1)} \end{cases} \quad (9)$$

Since the four nearest neighboring pixels along the diagonal directions within the R color plane have stable correlation with the expected value $\tilde{R}_{(i,j)}$, the obtained $\hat{R}_{(i,j)}$ is further refined by the following equation:

$$\begin{cases} \tilde{R}_{(i,j)} = \delta \cdot \hat{R}_{(i,j)} + (1-\delta) \cdot \sum \mu_k v_k / \sum v_k \\ \mu_k = R_k - \tilde{R}_{(i,j)}, \quad v_k = 1/1 + |\mu_k| \end{cases} \quad (10)$$

where $k$ represents four neighboring pixels. The factor $\delta$ is used to adjust the refinement performance. We set it experimentally as a constant value between 0.5 and 0.7 to obtain the optimal reconstruction. Using the nearest neighboring pixels $\tilde{R}$ and known R components along the four directions, the interpolation process for the missing R components at G sampling positions is identical to restoration of the missing R components at B sampling positions. The B channel can be demosaicked by means of the same procedure as R channel.

*Experimental results:* To evaluate the proposed method, we have conducted testing experiments and compared its performance against four state-of-the-art algorithms: DLMMSE [2], MSG [4], MDWI [5] and MLRI+wei [6]. The representative images are selected from the Kodak and IMAX datasets. These full color images are first decomposed into monochrome images with the Bayer pattern. The regularization parameter $\lambda$ is set as $2 \times 10^{-3}$. The spatial step $h$ is set to 5. $r_1$, $r_2$, $r_3$ and $r_4$ are set as 40, 40, 100 and 100, respectively. The stopping condition for iteratively solving Eq. (5) is $\|\tilde{G}^k - \tilde{G}^{k-1}\|_2 / \|\tilde{G}^k\|_2 \leq 10^{-4}$. The factor $\delta$ is selected as 0.6.

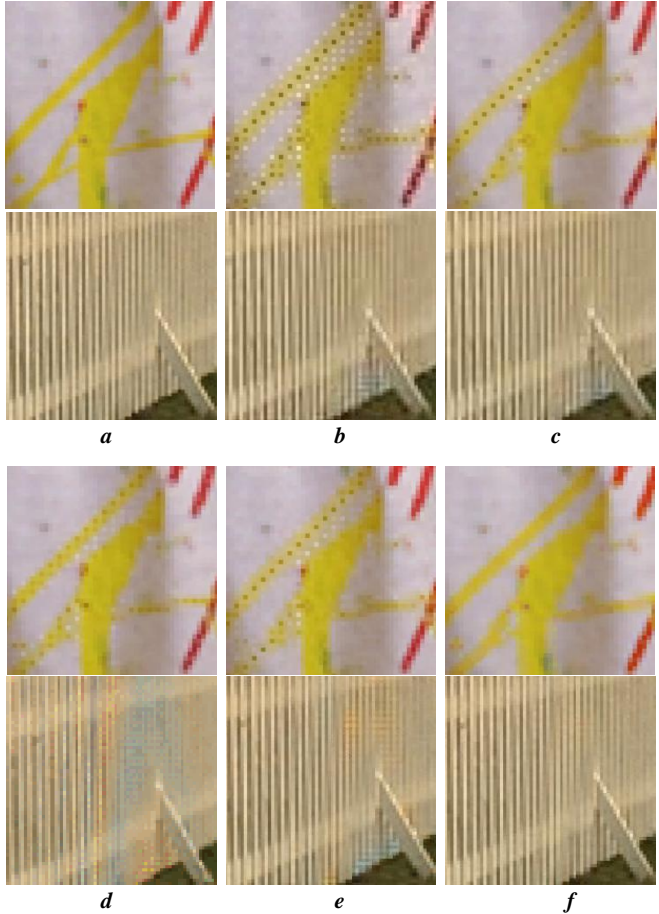

**Fig. 2** *Visual comparison of different demoaicking methods*
*a* Cropped regions of the original images
*b* DLMMSE
*c* MSG
*d* MDWI
*e* MLRI+wei
*f* Proposed

From demosaicked images shown in Fig. 2, we can observe that compared with other demosaicking algorithms, the Bayer images reconstructed by the proposed method have much higher subjective quality and less color artifacts, which involve blocking, blurring and zipper effect at edges. To measure the demosaicking quality, the average S-CIELAB and CPSNR values [6] are computed on two image datasets. It can be seen from Table 1 that our method has totally achieved higher reconstruction performance. This is mainly because our MC models are very effective to reduce color artifacts and much less sensitive to the change of spectral correlation.

**Table 1:** Average S-CIELAB and CPSNR (in dB) results

| Datasets | IMAX | | Kodak | |
|---|---|---|---|---|
| | S-CIELAB | CPSNR | S-CIELAB | CPSNR |
| DLMMSE | 1.436 | 34.46 | 1.309 | 39.58 |
| MSG | 1.368 | 34.59 | 1.162 | 41.00 |
| MDWI | 1.081 | 36.42 | 0.717 | 41.96 |
| MLRI+wei | 1.035 | 36.91 | 0.624 | 42.74 |
| **Proposed** | **1.020** | **37.23** | **0.586** | **43.15** |

*Conclusion:* In this Letter, we propose an efficient demosaicking algorithm by developing two optimized MC models to enhance the interpolation accuracy and suppress color artifacts. By exploiting the mean curvature of G image surface as a regularizer, the variational MC model is iteratively solved to reconstruct full-resolution G image with high accuracy and powerful structure-preserving capability. Moreover, this model can be incorporated into other demosaicking methods. Based on high correlation in curvature domain, the R and B full-resolution images can be well restored using the linear MC model and a postprocessing refinement. Extensive experiments have fully showed the superior performance of our method.

*Acknowledgments:* This work is supported by China Postdoctoral Science Foundation (2014M560852) and Major National Scientific Instrument and Equipment Development Project of China (2013YQ030967).

Rui Chen, Huizhu Jia, Xiange Wen and Xiaodong Xie (*National Engineering Laboratory for Video Technology*, *Peking University, Beijing 100871, People's Republic of China*)
E-mail: hzjia@pku.edu.cn